\documentclass[a4paper,12pt]{article}

\usepackage{a4,a4wide}
\usepackage{times}
\usepackage{latexsym}
\usepackage{amstext,amssymb}
\usepackage{named}

\newcommand{\BD}{\text{BD}}

\newcommand{\wfds}{\text{WFDS}}
\newcommand{\wfdh}{\text{WFDH}}
\newcommand{\wfs}{\text{WFS}}

\newcommand{\dwfs}{\mbox{D-WFS}}
\newcommand{\uwfs}{\text{U-WFS}}

\newcommand{\aeb}{\text{AEB}}

\newcommand{\static}{\text{STATIC}}
\newcommand{\dsls}{\text{D-SLS}}
\newcommand{\texta}{A}
\newcommand{\textb}{B}
\newcommand{\textc}{C}
\newcommand{\cons}{\text{cons}}
\newcommand{\dwfsnew}{\text{D-WFS}^*}
\newcommand{\wfdsnew}{\text{WFDS}}
\newcommand{\Lft}{\text{Lft}}
\newcommand{\res}{\text{res}}
\newcommand{\DB}{\text{DB}}
\newcommand{\lfp}{\text{lfp}}
\newcommand{\GDB}{\text{GDB}}

\newcommand{\wrt}[0]{w.r.t.\ }
\newtheorem{proposition}{Proposition}[section]
\newtheorem{example}{Example}[section]
\newcommand{\D}{\;|\;}

\newtheorem{definition}{Definition}
\newtheorem{theorem}{Theorem}[section]
\newtheorem{corollary}{Corollary}[section]
\newtheorem{lemma}{Lemma}[section]

\newcommand{\head}[1]{\mathit{head}(#1)}
\newcommand{\pbody}[1]{\mathit{body}^+(#1)}
\newcommand{\nbody}[1]{\mathit{body}^-(#1)}
\newcommand{\body}[1]{\mathit{body}(#1)}

\newcommand{\nafo}[0]{\mathit{not}}
\newcommand{\naf}[0]{\nafo\;}

\newcommand{\la}{\leftarrow}

\newcommand{\ms}{\text{ms}}
\newcommand{\can}{\text{can}}

\newcommand{\Sem}{\text{Sem}}

\title{Comparisons and Computation of Well-founded Semantics for Disjunctive 
       Logic Programs}
\author{Kewen Wang\thanks{\ This work was done while the
                      second author was with the 
                      University of Potsdam.}
            \\
School of Computing and Information Technology\\
Griffith University, QLD 4111, Australia \\
\tt{k.wang@cit.gu.edu.au} \\
\tt{http://www.cit.gu.edu.au/$\sim$s2107085/} \\
Lizhu Zhou \\
Department of Computer Science and Technology\\
Tsinghua University, Beijing\\
\tt{dcszlz@tsinghua.edu.cn}
  }

\begin{document}

\date{}

\maketitle
\begin{abstract}
Much work has been done on 
extending the well-founded semantics to general disjunctive logic 
programs and various approaches have been proposed. 
However, these semantics are different from each other and
no consensus is reached about which semantics is the most intended.
In this paper we look at disjunctive well-founded
reasoning from different angles.
We show that there is an intuitive form of
the well-founded reasoning in disjunctive logic programming which
can be characterized by slightly modifying some exisitng approaches to
defining disjunctive well-founded semantics,
including program transformations,
argumentation, unfounded sets (and resolution-like procedure).
We also provide a bottom-up procedure for this semantics.
The significance of our
work is not only in clarifying the relationship among different approaches,
but also shed some light on what is an intended well-founded semantics
for disjunctive logic programs.
\end{abstract}
\section{Introduction}\label{section:introduction}
The importance of representing and reasoning about disjunctive
information has been addressed by many researchers. 
As pointed out in \cite{bradix99}, the related application domains include
reasoning by cases, legal reasoning, diagnosis, natural language
understanding and conflict resolving in multiple inheritance.
Disjunctive logic programming (DLP) is widely believed to be
a suitable tool for formalizing disjunctive reasoning and 
it has received extensive studies in recent years, 
e.~g. \cite{aptbol94,bargel94,gellif91,lomira92}.
Since DLP  admits both default negation and disjunction,
the issue of finding a suitable semantics for disjunctive programs
is more difficult than it is in the case of
normal (i.~e. non-disjunctive) logic programs.
Usually, skepticism and credulism represent two major
semantic intuitions for knowledge representation in artificial intelligence.
The well-founded semantics \cite{varosc91} is a formalism 
of skeptical reasoning in normal logic programming while
the stable semantics \cite{gellif88} formalizes credulous
reasoning. 
Recently, considerable effort has been paid
to generalize these two semantics to disjunctive logic programs.
However, the task of generalizing the well-founded model to
disjunctive programs has been proven to be complex.
There have been various proposals for defining the well-founded semantics
for general disjunctive logic programs \cite{lomira92}.
As argued by some authors (for instance \cite{bradix99,prz95,wan00}),
each of the previous versions of the disjunctive well-founded semantics
bears its own drawbacks. 
Moreover,
no consensus is reached about what constitutes an intended well-founded
semantics for disjunctive logic programs.
The semantics {\dwfs} \cite{bradix98,bradix99}, STATIC \cite{prz95}
and WFDS \cite{wan00}
are among the most recent approaches to defining disjunctive
well-founded semantics. 
{\dwfs} is based on a series of abstract properties
and it is the weakest (least) semantics that is invariant under
a set of program transformations.
STATIC has its root in autoepistemic logic and is based
on the notion of {\em static expansions} for belief theories.
The semantics $\static (P)$ for a disjunctive program $P$
is defined as the least static expansion
of $P_{\aeb}$ where $P_{\aeb}$ is the belief theory corresponding to $P$.
The basic idea of WFDS is to transform
$P$ into an argumentation framework and 
$\wfds (P)$ is specified by the least acceptable hypothesis of $P$.
Although these semantics stem from very different intuitions,
all of them share a number of attractive properties. 
For instance, each of these semantics extends both
the well-founded semantics~\cite{varosc91} for normal logic programs
and the generalized closed world assumption (GCWA)~\cite{min82}
for positive disjunctive programs (i.~e. without default negation);
each of these semantics is consistent and provides approximation to
the disjunctive stable semantics (i.e. a literal derived under the well-founded
semantics is also derivable from any stable model).

However, the problem of comparing different approaches to defining
disjunctive well-founded semantics is rarely investigated.
A good starting point is \cite{brdinipr01} in which it is proven that 
$\dwfs$ is equivalent to a restricted version of STATIC. 
But the relation of $\dwfs$ to the argumentation-based semantics
and unfounded sets is as yet unclear. More importantly,
it is an open question whether there is a disjunctive well-founded semantics
that can be characterized by all of these approaches.
 
In this paper, we intuitively (and slightly) modify some existing semantics
and report further equivalence results:
\begin{enumerate}
\item As we will see in Section~\ref{section:trans}, the transformation-based
semantics $\dwfs$ is different from $\wfds$ and seems a little too skeptical. 
The reason is that the program transformations in Brass and Dix's set ${\bf T}_{\wfs}$
are unable to reduce the rule head if we do not remove a rule from the
disjunctive program. Interestingly, this problem is related to
the famous GCWA (Generalized Closed World Assumption) \cite{min82}. 
Based on this observation, we introduce a new program transformation
called the {\em Elimination of s-implications}, which naturally extends the
Elimination of nonminimal rules in~\cite{bradix99}.

We define a new transformation-based semantics, denoted $\dwfsnew$,
as the weakest semantics that allows the Elimination of s-implications and
the program transformations in ${\bf T}_{\wfs}$ except for the
Elimination of nonminimal rules. 
This semantics naturally extends $\dwfs$ and enjoys all the important 
properties that have been proven for $\dwfs$.
An important result in this paper is that $\wfds$ is equivalent to $\dwfsnew$,
which establishes a precise relationship between argumentation-based approach
and transformation-based approach in disjunctive logic programming.

\item The notion of {\em unfounded sets} is well-known in logic programming.
It was first employed by \cite{varosc91} to define the well-founded semantics
for normal logic programs and then to characterize other semantics including
the stable models \cite{gellif88} and partial stable models \cite{saczan90}.
This notion has been generalized from normal to disjunctive logic programs
\cite{eilesa97,lerusc97}. Although a form of disjunctive well-founded
semantics is investigated
in \cite{lerusc97}, the generalized unfounded sets are mainly used to
characterize partial stable models for disjunctive programs.
Moreover, their notions are not appropriate for defining a disjunctive
well-founded semantics as we will see in Section~\ref{section:unfounded}.
One reason for this is that their notions are defined only for interpretations
(i.~e. consistent sets of literals) rather than for model states (i.~e. sets 
of disjunctions of literals) \cite{lomira92}. 
Thus, we further generalize the notion of unfounded sets to model states.
The resulting disjunctive well-founded semantics, denoted $\uwfs$, behaves more intuitive
and actually we show that it is also equivalent to $\wfds$.

\item
We develop a bottom-up evaluation procedure for $\wfds$ (equivalently, 
for $\dwfsnew$, $\uwfs$) in a similar way as in \cite{bradix99}.
Specifically, for each disjunctive program $P$, it can be gradually transformed into
a normal form called {\em strong residual program} $\res^*(P)$ by
our elementary program transformations. 
We show that the semantics $\wfds(P)$ can be directly read out from 
the strong residual program: if there is no rule head containing
an atom $p$, then $\naf p\in\wfds(P)$; if there is a rule of the form
$\texta\leftarrow$ in the strong residual program, then
$\texta\in\wfds(P)$.
That is, our bottom procedure is sound and complete with respect to
$\wfds$.
\end{enumerate}

Moreover, in \cite{wan01b} we have developed a top-down procedure D-SLS Resolution
which is sound and complete with respect to $\wfds$.
D-SLS naturally extends both SLS-resolution~\cite{ros92}
(for normal logic program) and 
SLI-resolution~\cite{lomira92} (for disjunctive programs without default negation). 

Altogether we obtain the following equivalence results:   
\[
\wfds \equiv \dwfsnew  \equiv \uwfs \equiv \dsls.
\]
We consider these results to be quite significant:
\begin{enumerate}
\item Our results clarify the relationship among several
different approaches to defining disjunctive well-founded semantics,
including argumentation-based, transformation-based, unfounded sets-based
(and resolution-based approaches).
\item Since the four semantics are based on very
different intuitions, these equivalent characterizations 
in turn shed some light on what is an intended well-founded semantics for
disjunctive logic programs.
\item The bottom-up query evaluation proposed in this paper paves a promising way
for implementing disjunctive well-founded semantics. 
\end{enumerate}
The rest of this paper is arranged as follows.
In Section~\ref{section:preliminary} we recall some basic definitions and notation;
we present in Section~\ref{section:argu} a slightly restricted form of the well-founded 
semantics $\wfds$ (we still denote $\wfds$). 
In Section~\ref{section:trans} we introduce the program transformation \emph{
Elimination of s-implications} and then define the 
transformation-based semantics $\dwfsnew$, which naturally extends $\dwfs$. 
In Section~\ref{section:coindence}, we first provide a bottom-up query evaluation
for $\dwfsnew$ (equivalently, $\wfds$) and then prove the equivalence of 
$\dwfsnew$ and $\wfds$.
Section~\ref{section:unfounded} introduces a new notion of unfounded sets
and defines the well-founded semantics $\uwfs$. We also show that
$\uwfs$ is equivalent to $\wfds$. Section~\ref{section:conclusion} is our
conclusion. 

\section{Preliminaries}\label{section:preliminary}
We briefly review most of
the basic notions used throughout this paper.

A {\em disjunctive logic program} is a finite set of rules  of the form
\begin{equation}\label{rule}
a_1\vee \cdots \vee a_n\leftarrow 
       b_1,\ldots,b_m,\naf c_1,\ldots,\naf c_t,
\end{equation}
where $a_i,b_i,c_i$ are atoms and $n> 0$.
The default negation `$\naf a$' of an atom $a$ is
called a {\em negative literal}.
 
In this paper we consider only propositional programs although
some of the definitions and results hold for predicate logic programs.

For technical reasons, it should be stressed that the body of a rule is a set of 
literals rather than a multiset. For instance, $a\vee b\leftarrow c, c$ is not a rule in our sense
while $a\vee b\leftarrow c$ is a rule. 
That is, we assume that any rule of a logic program has been
simplified by eliminating repeated literals in both its head and body.

$P$ is a {\em normal} logic program if it contains
no disjunctions.

If a rule of form (\ref{rule}) contains no negative body literals,
it is called \emph{positive}; $P$ is a positive program if every rule
of $P$ is positive.

If a rule of form (\ref{rule}) contains no body atoms,
it is called \emph{negative}; $P$ is a negative program if every rule
of $P$ is negative.

Following~\cite{bradix99}, we also say a negative rule $r$ is a
\emph{conditional fact}.
That is, a conditional fact is of form
$a_1\vee\cdots\vee a_n \la \naf c_1,\cdots,\naf c_t$,
where $a_i$ and $c_j$ are (ground) atoms for $1\le k\le n$ and 
$0\le j \le t$.

For a rule $r$ of form (\ref{rule}), 
$\body{r}=\pbody{r}\cup \nbody{r}$
where $\pbody{r}=\{b_1,\ldots,b_m\}$
and $\nbody{r}=\{\naf c_1,\ldots,\naf c_t\}$; 
$\head{r}=a_1\vee \cdots \vee a_n$. When no confusion is caused,
we also use $\head{r}$ to denote the set of atoms in $\head{r}$.
For instance,  $a\in \head{r}$ means that $a$ appears in the head of $r$.
If $X$ is a set of atoms, $\head{r}-X$ is the disjunction
obtained from $\head{r}$ by deleting the atoms in $X$.
The set $\head{P}$ consists of all atoms appearing
in rule heads of $P$.

In the sequel, we will use the capital letters $A,B,C$ to
represent both disjunctions or sets of atoms (in case there is confusion,
we will explicitly claim their scopes).

As usual, $B_P$ is the {\em Herbrand base} of disjunctive 
logic program $P$, that is,
the set of all (ground) atoms in $P$. A {\em positive}
({\em negative}) 
disjunction is a disjunction of atoms (negative literals) of $P$.
A {\em pure disjunction} is either a positive one or a negative one. 

The disjunctive base of $P$ is ${\DB}_P={\DB}_P^+\cup {\DB}_P^-$ where
${\DB}_P^+$ is the set of all positive disjunctions in $P$ and 
${\DB}_P^-$ is the set of  all negative disjunctions in $P$.
If $\texta$ and $\textb=\texta\vee \texta'$ are two disjunctions, 
then we say
$\texta$ is a {\em sub-disjunction} of $\textb$, 
denoted $\texta\subseteq \textb$. $\texta\subset \textb$ means
$\texta\subseteq \textb$ but $\texta\neq \textb$

A {\em model state} of a disjunctive program $P$ is a subset of ${\DB}_P$.

A model state $S$ is {\em inconsistent} if at least one of the following two
conditions holds:
\begin{enumerate}
\item There is a positive disjunction
$a_1\vee\cdots\vee a_n\in S$ ($n\ge 1$) such that $\naf a_i\in S$ for all
$1\le i\le n$; 
or
\item There is a negative disjunction
$\naf a_1\vee\cdots\vee \naf a_n\in S$ ($n\ge 1$) such that $a_i\in S$ for all
$1\le i\le n$.
\end{enumerate}
Otherwise, we say that $S$ is a {\em consistent} model state.

Usually, a well-founded semantics for a disjunctive logic program is defined
by a (consistent) model state.

If $E$ is an expression (a set of literals, a disjunction or 
a set of disjunctions), 
$atoms(E)$ denotes the set of all atoms appearing in $E$.

For simplicity, we assume that all model states are closed under
implication of pure disjunctions. That is, for any model state $S$,
if $A$ is a sub-disjunction of a pure disjunction $B$ and
$A\in S$, then $B\in S$. For instance, if $S=\{a, b\vee c\}$,
then we implicitly assume that $a\vee b\vee c \in S$.

Given a model state $S$ and a pure disjunction $A$, we also say $A$
is satisfied by $S$, denoted $S\models A$, if $A\in S$.

We assume that all disjunctions have been simplified by deleting
the repeated literals. For example, the disjunction $a\vee b\vee b$
is actually the disjunction $a\vee b$.

For any set $S$ of disjunctions, the {\em canonical form} of $S$ is defined as 
$can(S)=\{A\in S \mid \text{there is no disjunction $A'\in S$ s.~t. }A'\subset A\}.$

We recall that the \emph{least model state} of a positive disjunctive
program $P$ is defined as
\[
\ms(P)=\{A\in\DB_P^+\D P\vdash A\}.
\]
Here $\vdash$ is the inference relation of the classical propositional logic.

Given a positive disjunctive program $P$, $\can(\ms(P))$ can also be
equivalently characterized by the least fixpoint of the immediate consequence
operator $T_P^G$  for $P$ (see \cite{lomira92} for details).
%
\begin{definition} \label{def:imediate:operator:original}
Let $P$ be a positive disjunctive program and let $J$ be a subset of $\DB_P^+$.
The immediate consequence operator
$T_P^S:\; 2^{\DB_P^+}\rightarrow 2^{\DB_P^+}$ is defined as follows
\[
T_P^S(J)
\;=\;
\left\{
  A\in\DB_P^+\left|\; 
    \begin{array}{rl}
           & \text{there exist a rule}\;
       A'\leftarrow b_1,\ldots,b_m
            \\
           & \text{in $P$ and } A_1, \ldots , A_m\in\DB_P^+\text{ such that}
            \\
      (1). & (b_i\vee A_i)\in J, \text{ for all }i=1, \ldots ,m; and
             \\
      (2). & A=A'\vee A_1\vee\cdots\vee A_m
    \end{array}
  \right\}\right.
\]
\end{definition}
%
$T_P^S$ is actually the following hyperresolusion:
\begin{equation}\label{eqn:hyperresolusion}
\frac{b_1\vee A_1;\; \ldots,b_m\vee A_m;\; A'\leftarrow b_1,\ldots,b_m}%
     {A'\vee A_1\vee\cdots\vee A_m}
\end{equation}
Note that we always remove repetitions of literals in each rule and/or
disjunctions.

Define $T_P^S\uparrow 0=\emptyset$, and
$T_P^S\uparrow (n+1)=T_P^S(T_P^S\uparrow n)$
for $n\ge 0$. Then we have the following result~\cite{lomira92}.
%
\begin{theorem} \label{thm:can:ms}
Let $P$ be a positive disjunctive program. Then,

$\can(\ms(P))=can(\lfp(T_P^S))=can(T_P^S\uparrow\omega)$.
\end{theorem}
%

\section{Argumentation and well-founded semantics}
\label{section:argu}
As illustrated in \cite{wan00}\footnote{You et al in~\cite{yoyugo00} also defined
an argumentative extension to the disjunctive stable semantics. However, their framework
does not lead to an intuitive well-founded semantics for DLP
as the authors have observed.}, argumentation provides
an unifying semantic framework for DLP.
The basic idea of the argumentation-based approach for DLP  is
to translate each disjunctive logic program into an
argument framework 
${\bf F}_P=\langle P, {\DB}_P^-,{\leadsto}_P\rangle$.
In that framework, an {\em assumption} of $P$
is a negative disjunction of $P$, and a {\em hypothesis}
is a set of assumptions; 
${\leadsto}_P$ is an attack relation 
among the hypotheses.
An {\em admissible hypothesis} $\Delta$ is one that can attack 
every hypothesis which attacks it.
The intuitive meaning of an assumption 
$\naf a_1\vee\cdots\vee \naf a_m$
is that $a_1\wedge\cdots\wedge a_m$ can not be proven
from the disjunctive program.

Given a hypothesis $\Delta$ of disjunctive program $P$, 
similar to the GL-transformation \cite{gellif88},
we can easily reduce $P$ into another disjunctive program
without default negation. 
%
\begin{definition}\label{def:reduct}
Let $\Delta$ be a hypothesis of disjunctive program $P$, then
the reduct of $P$ with respect to $\Delta$ is the
disjunctive program
\[
P^+_\Delta =  \{\head{r}\la \pbody{r} \D  r\in P
                            \text{ and }\nbody{r}\subseteq \Delta\}.
\]
\end{definition}
%
It is obvious that $P^+_\Delta$ is a positive disjunctive program
(i~.e. without default negation).
%
\begin{example}
Let $P$ be the following disjunctive program:
\[
\begin{array}{rcl}
a            &\la &b,\naf c\\
b\vee c      &\la &\naf e\\
b\vee c\vee d&\la &
\end{array}
\]
If $\Delta=\{\naf c\}$, then $P_{\Delta}^+$ is the positive program:
\[
\begin{array}{rcl}
a            &\la &b\\
b\vee c\vee d&\la &
\end{array}
\]

Thus, $\can(ms(P_{\Delta}^+))=\{a\vee c\vee d,b\vee c\vee d\}$.
\end{example}
%
The following definition introduces a special resolution ${\vdash}_P$ 
which resolves default-negation literals with a disjunction.
%
\begin{definition} \label{def:vdash_P}
Let $\Delta$ be a hypothesis of disjunctive program
$P$ and $\texta\in {\DB}_P^+$.
If there exist $\textb\in {\DB}_P^+$ and $\naf b_1,\ldots,
\naf b_m\in \Delta$ such that 
$\textb =\texta\vee b_1\vee \cdots \vee b_m$ and
$\textb\in\can(\ms(P_{\Delta}^+))$.
Then $\Delta$ is said to be a {\em supporting hypothesis} 
for $\texta$, denoted $\Delta {\vdash}_P \texta$. 

The set of all positive disjunctions supported by $\Delta$ is denoted:
\[
{\cons}_P(\Delta)=\{\texta \in {\DB}_P^+ \D \Delta {\vdash}_P \texta\}. 
\]
\end{definition}
%
%
\begin{example}\label{exa:vdash_P}
Consider the following disjunctive program $P$:
\[
\begin{array}{rcl}
a\vee b & \la & c,\naf d\\
c\vee e & \la & g, \naf f\\
a\vee d & \la & \naf b \\
g       & \la & 
\end{array}
\]
Let $\Delta=\{\naf e, \naf d, \naf f\}$.
Then $P^+_\Delta$ consists of the following
three rules:
\[
\begin{array}{rcl}
a\vee b & \la & c\\
c\vee e & \la & g\\
g       & \la & 
\end{array}
\]
Thus, $\can(\ms(P_{\Delta}^+))=
\{g,c\vee e,a\vee b\vee e\}$.
Since $\naf e$ is in $\Delta$, we have
$\Delta\vdash_P a\vee b$.
\end{example}
To specify what is an acceptable hypothesis for a given disjunctive program,  
some more constraints will be required so that unacceptable hypotheses
are ruled out.
%
\begin{definition}\label{def:attack}
Let $\Delta$ and ${\Delta}'$ be two hypotheses of disjunctive
program $P$. 
We say $\Delta$  {\em attacks} ${\Delta}'$, denoted
$\Delta \leadsto_P {\Delta}'$,
if at least one of the  following two conditions holds:

1.  there exists $\beta =\naf b_1\vee \cdots \vee \naf b_m \in {\Delta}',\;m>0,$ such that
$\Delta {\vdash}_P b_i$, for all $i=1,\ldots,m$; or

2.  there exist $\naf b_1,\ldots,\naf b_m \in {\Delta}',m>0$, such that
$\Delta {\vdash}_P b_1\vee \cdots \vee b_m$,
\end{definition}
%
If $\Delta \leadsto_P {\Delta}'$, we also say $\Delta$ is an
attacker of $\Delta'$.
In particular, if $\Delta \leadsto_P\{\naf p\}$, we simply say
that $\Delta$ is an attacker of the assumption $\naf p$.

Intuitively,  $\Delta \leadsto_P {\Delta}'$  means that $\Delta$ causes
a direct contradiction with $\Delta'$ and the contradiction may come
from one of the two cases in Definition~\ref{def:attack}. 
\begin{example}\label{exa:dix:mod}
\[
\begin{array}{rcl}
a\vee b & \la & \\
c       & \la & d, \naf a, \naf b\\
d       & \la & \\
e       & \la & \naf e 
\end{array}
\]
Let $\Delta'=\{\naf c\}$ and $\Delta=\{\naf a, \naf b\}$, then
$\Delta \leadsto_P {\Delta}'$.
\end{example}
%
The next definition defines what is an acceptable hypothesis.
\begin{definition}\label{def:acc}
 Let $\Delta$ be a hypothesis of disjunctive program $P$. 
An assumption $\textb$ of $P$
is {\em admissible} with respect to $\Delta$ if
$\Delta{\leadsto}_P {\Delta}'$ holds for any hypothesis ${\Delta}'$ 
of $P$ such that $\Delta'\leadsto_P\{\textb\}$. 

Denote 
\({\mathcal A}_P(\Delta)=\{\naf a_1\vee\cdots\vee \naf a_m \in {\DB}_P^-\D
\naf a_i
\text{ is admissible wrt}
\)
\(
\Delta \text{ for some }
i, 1\le i\le m\}.
\)
\end{definition}
Originally, ${\mathcal A}_P$ also includes some other negative disjunctions.
To compare with different semantics, we omit them here. Another reason for
doing this is that information in form of negative disjunctions
does not participate in inferring positive information in DLP.
 
For any disjunctive program $P$,
${\mathcal A}_P$ is a monotonic operator.
Thus ${\mathcal A}_P$ has the least fixpoint $\lfp({\mathcal A}_P)$ and 
$\lfp({\mathcal A}_P)={\mathcal A}^k_P(\emptyset)$ for some $k\ge 0$
if $P$ is a finite propositional program.
\begin{definition}
The well-founded disjunctive hypothesis ${\wfdh}(P)$ 
of disjunctive program $P$ is defined as
the least fixpoint of the operator ${\mathcal A}_P$. That is,
${\wfdh}(P)={\mathcal A}_P\uparrow \gamma$, where $\gamma$ is an ordinal.

The well-founded disjunctive semantics $\wfdsnew$ for $P$ is defined as
the model state $\wfdsnew (P)={\wfdh}(P)\cup {\cons}_P({\wfdh}(P))$.
\end{definition}
%
By the above definition, $\wfdsnew (P)$ is uniquely determined by
${\wfdh}(P)$.

For the disjunctive program $P$ in Example~\ref{exa:dix:mod},
$\wfdh(P)=\{\naf c\}$ and $\wfdsnew(P)=\{a\vee b, d, \naf c\}$.
Notice that $e$ is unknown.

A plausible hypothesis should not attack itself.
%
\begin{definition}
A hypothesis $\Delta$ is self-attacking if
$\Delta \leadsto_P \Delta$.
Otherwise, we say $\Delta$ is self-consistent.
\end{definition}
%
It has been proven in~\cite{wan00} that $\wfds$ is consistent
in the following sense.
%
\begin{theorem} \label{thm:wfds:consistent}
For any disjunctive program $P$,
$\wfdh(P)$ is self-consistent and thus
$\wfds(P)$ is a consistent model state.
\end{theorem}
%
\section{An Alternative Definition of \wfds}
\label{sec:alternative}
There are several alternative ways of defining argumentative semantics
for disjunctive programs and this issue is often confused in literature.
In this section, we will try to explain why some of the possible
alternatives are unintuitive and then provide an equivalent definition
for \wfds. 

One may ask why we cannot replace the inference relation $\vdash_P$
with the classical inference relation. This can be clearly explained
by the following example. 
Let $\wfds_1$ denote the disjunctive well-founded semantics
obtained by replacing the inference relation $\vdash_P$
in Definition~\ref{def:vdash_P} with the classical inference relation $\vdash$.
%
\begin{example}
Let $P$ be the following logic program:
\[
\begin{array}{rcl}
a  &\la &\naf b\\
c  &\la &\naf c
\end{array}
\]
For this program $P$, its intuitive semantics should be $M_0=\{a, \naf b\}$.
That is, $a$ is true and $b$ is false while $c$ is undefined.
In fact, the well-founded semantics for non-disjunctive logic program
assigns the model $M_0$ to $P$.

However, if we replace $\vdash_P$ with the classical inference relation
$\vdash$ in Definition~\ref{def:vdash_P}, then the resulted disjunctive
well-founded semantics $\wfds_1$ will derive nothing from $P$, i.~e.
$a,b,c$ will be all undefined (since $\dwfs_1=\emptyset$). 
To see this, let $\Delta'=\{\naf c\}$ and then
$P^+_{\Delta'}\vdash c$. So, $P^+_{\Delta'}\vdash c\vee b$.
Since $\naf c\in\Delta'$, we have
$P^+_{\Delta'}\vdash b$. This means $\{\naf c\}$ is an attacker
of the assumption $\naf b$. However, $\emptyset$ cannot attack
$\{\naf c\}$.
\end{example}
%
One might further argue that the unintuitive behavior above of $\wfds_1$
is not caused by replacing $\vdash_P$ with the classical inference relation
$\vdash$ in Definition~\ref{def:vdash_P} but by our allowing
self-attacking hypothesis $\Delta'=\{\naf c\}$.
So, we might try to require that the attacker $\Delta'$ in
Definition~\ref{def:acc} is self-consistent and denote the resulted
semantics as $\dwfs_2$. 
This modification causes an unintended semantics again.
For example, let $P$ consist of only one rule $a\vee c\la \naf c$.
Although $\{\naf c\}\leadsto_P a$ but $\{\naf c\}$ is self-attacking.
That is, the assumption $\naf a$ has no self-consistent attacker
and thus $\naf a\in\wfds_2$. This result contradicts to all of the
existing well-founded semantics for disjunctive programs.

We have another possibility of modifying
Definition~\ref{def:vdash_P}. 
Specifically, we can replace $\can(\ms(P^+_\Delta))$
with $T^S_{P^+_\Delta}\uparrow\omega$ and the resulted inference relation 
is denoted
as $\vdash'_P$. Parallel to Definition~\ref{def:acc}, we can define
a new attack relation $\leadsto'_P$ and thus a new 
disjunctive well-founded semantics denoted $\dwfs'$.

The inference $\vdash'_P$ looks more intuitive than $\vdash_P$
and in fact we will provide a resolution-like definition for it
in the following.

Notice that the inference relation $\vdash'_P$ is actually a combination
of the following two inference rules (the first one is a generalization
of the SLI-resolution~\cite{lomira92}):

\begin{equation}\label{rule:sli}
\frac{A\la a,B,\naf C;\quad a\vee A'\la\naf C'}%
     {A\vee A'\la B,\naf C,\naf C'}
\end{equation}

\begin{equation}\label{rule:naf:reso}
\frac{p_1\vee\cdots\vee p_s\vee A\leftarrow\naf C;
      \quad \naf C\cup\{\naf p_1,\ldots,\naf p_s\}}{A}
\end{equation}
Here $A$ is a positive disjunction. 

The intuition of rule~(\ref{rule:naf:reso}) is quite simple:
If we have the hypothesis $\naf C\cup\{\naf p_1,\ldots,\naf p_s\}$, 
then we can infer $A$ from the program rule 
$p_1\vee\cdots\vee p_s\vee A\la \naf C$.
Moreover, we can fully perform the rule~(\ref{rule:sli}) in advance and then 
apply the rule~(\ref{rule:naf:reso}).

Since $\can(\ms(P^+_\Delta))\subseteq T^S_{P^+_\Delta}\uparrow\omega$,
we have $\Delta\vdash'_P A$ implies $\Delta\vdash_P A$
for any hypothesis $\Delta$ and any disjunction $A$.
However,
$\vdash'_P$ is different from $\vdash_P$ in general.
For instance, let $P=\{a\la; \quad a\vee b\la\}$ and $\Delta=\emptyset$.
Then $T^S_{P^+_\Delta}\uparrow\omega=\{a, a\vee b\}$ while
$\can(\ms(P^+_\Delta))=\can(T^S_{P^+_\Delta}\uparrow\omega)=\{a\}$.
Even if this fact, the disjunctive well-founded semantics based on
these two inference relations become equivalent.

The main result of this section is thus the equivalence of $\wfds$ and
$\wfds'$.
%
\begin{theorem}\label{thm:wfds:alt}
For any disjunctive program $P$, we have
\[
\wfds'(P)=\wfds(P).
\]
\end{theorem}
%
Having this theorem, we will be able to use $\wfds$ to denote
both $\wfds'$ and $\wfds$ in the following sections.
%
\begin{proof}{thm:wfds:alt}
Denote the set of admissible hypotheses for $\Delta$ wrt $\wfds'$
as ${\mathcal A}'_P(\Delta)$ where $P$ is a disjunctive program
and $\Delta$ is any hypothesis of $P$.

It suffices to show that
\[
{\mathcal A}_P\uparrow\omega
=
{\mathcal A}'_P\uparrow\omega.
\]
This is further reduced to proving that
\[
{\mathcal A}_P\uparrow k
=
{\mathcal A}'_P\uparrow k
\]
for all $k\ge 0$.

We use induction on $k$.

For simplicity, write
\(
\Delta_k={\mathcal A}_P\uparrow k
\)
and 
\(
\Delta'_k={\mathcal A}'_P\uparrow k.
\)

For $k=0$, it is obvious since $\Delta_0=\Delta'_0=\emptyset$.

Assume $\Delta_k=\Delta'_k$, we want to show that
$\Delta_{k+1}=\Delta'_{k+1}$.

If $\naf p\in \Delta_{k+1}$, then $\Delta_k\leadsto_P\Delta'$ for
any hypothesis $\Delta'$ with $\Delta'\vdash_P p$.

For any hypothesis $\Delta'$, if $\Delta'\vdash'_P p$,
consider two possible cases:

Case 1.\quad $\Delta'\vdash_P p$: then 
$\Delta_k\leadsto_P\Delta'$ and thus
$\Delta_k\leadsto'_P\Delta'$.

Case 2.\quad $\Delta'\not\vdash_P p$: then
the following conditions are satisfied:

(1).  $p\not\in\can(\ms(P^+_\Delta))$; and

(2).  There is a disjunction $A=a_1\vee\cdots\vee a_m$ such that
$\{\naf a_1,\ldots,\naf a_m\}\subseteq\Delta'$,
$p\vee A\in T^S_{P^+_{\Delta'}}\uparrow\omega$.

By the above two conditions, there is a sub-disjunction $A'$ of
$A$ such that $A'\in\can(\ms(P^+_\Delta))$ and $p\not\in A'$.
Thus, $\emptyset\leadsto_P\Delta'$.
This implies $\Delta\leadsto_P\Delta'$.
So, $\Delta\leadsto'_P\Delta'$.
That is, $\naf p\in \Delta'_{k+1}$.

For the opposite direction, suppose $\naf p\in\Delta'_{k+1}$.
Then $\Delta'_k\leadsto'_P\Delta'$ for any hypothesis $\Delta'$
with $\Delta'\vdash'_P p$.

For any hypothesis $\Delta'$, if $\Delta'\vdash_P p$,
then $\Delta'\vdash'_P p$, which implies
$\Delta_k\leadsto_P'\Delta'$ by $\naf p\in\Delta'_k$
and the induction assumption.
Consider two possible cases:

Case 1. There is an assumption 
$\naf a_1\vee\cdots\vee\naf a_m\in\Delta'$ such that
\(
\Delta_k\vdash'_P a_i
\)
for $1\le i\le m$:
Then for each $a_i$, there is a disjunction 
\(
a_i\vee b_1\vee\cdots\vee b_n\in T^S_{P^+_{\Delta_k}}
\)
such that $\{\naf b_1,\ldots,\naf b_n\}\subseteq\Delta_k$.
If 
\(
\Delta_k\not\vdash_P a_i
\)
for some $i (1\le i\le m)$, then there must be a
subdisjunction $B'$ of $b_1\vee\cdots\vee b_n$
such that
\(
B'\in\can(\ms(P^+_{\Delta_k}))
\).
This means $\Delta_k\leadsto_P\Delta_k$,
contradiction.

Therefore,
\(
\Delta_k\vdash_P a_i
\)
for $1\le i\le m$, which implies
$\Delta_k\leadsto_P\Delta'$.
Thus, $\naf p\in\Delta_{k+1}$.

Case 2. There are assumptions $\naf a_1,\ldots,\naf a_m\in\Delta'$
such that $\Delta'\vdash'_P a_1\vee\cdots\vee a_m$ where $m>0$
(without loss of generality, we can choose $m$ the least number):
Then there is a disjunction
\(
a_1\vee\cdots\vee a_m\vee b_1\vee\cdots\vee b_n
\in T^S_{P^+_{\Delta_k}}\uparrow\omega
\)
with $\naf b_1,\ldots,\naf b_n\in\Delta_k$.

On the contrary, suppose that
\(
\Delta_k\not\vdash_P a_i
\),
then  a
subdisjunction $B'$ of $b_1\vee\cdots\vee b_n$
is in
\(
\can(\ms(P^+_{\Delta_k}))
\)
by the minimality of $m$.
Thus, we also have $\Delta_k\leadsto\Delta_k$.
This means
\(
\Delta_k\vdash_P a_i
\)
and therefore,
$\naf p\in\Delta_{k+1}$.
\end{proof}
%

\section{Transformation-based semantics} \label{section:trans}
As mentioned in Section~\ref{section:introduction},
the transformation-based approach is a promising method
of studying semantics for DLP and based on this method,
a disjunctive well-founded semantics
called $\dwfs$ is defined in~\cite{bradix99}.
The authors first introduce some intuitive program transformations
and then define $\dwfs$ as the weakest semantics that satisfies their
transformations. In this section, we shall first analyze the insufficiency
of Brass and Dix's set of program transformations and then define a new  program transformation
called the \emph{Elimination of s-implications}, which is an
extension of a program transformation named 
the Elimination of nonminimal rules.
We then define a new transformation-based semantics, denoted $\dwfsnew$, as the
weakest semantics that allows the modified set of program transformations.
Our new semantics $\dwfsnew$ naturally extends
D-WFS and thus is no less skeptical
than D-WFS. More importantly, $\dwfsnew$ is equivalent to $\wfds$ as we will
show in Section~\ref{section:coindence}.

The primary motivation for extending $\dwfs$ is to define
a transformation-based counterpart for argumentation-based semantics.
However, this extension is also meaningful in view point of
commonsense reasoning,
because D-WFS seems too skeptical to derive useful information
from some disjunctive programs as the next example shows.
\begin{example}\label{exa:travel}
John is traveling in Europe but we are not sure which city he is visiting.
We know that, if there is no evidence to show that
John is in Paris, he should be either in
London or in Berlin. Also, we are informed that John is now visiting
either London or Paris. This knowledge base can be conveniently
expressed as the following disjunctive logic program $P_{\ref{eqn:travel}}$:
\begin{equation} \label{eqn:travel}
\begin{array}{crcl}
r_1: & b \vee l & \la & \naf p \\
r_2: & l \vee p & \la &
\end{array}
\end{equation}
Here, $b, l$ and $p$ denote that John is visiting {\em Berlin, London}
and {\em Paris}, respectively.

Intuitively, $\naf b$ (i.~e. John is not visiting Berlin) should be inferred
from $P$. It can be verified that neither $b$ nor its negation 
$\naf b$ can be derived from $P$ under
{\dwfs} or {\static} while $\naf b$ can be derived under \wfds.
\end{example}

Our analysis shows that this unwanted behavior of $\dwfs$ is caused by a
program transformation called \emph{the Elimination of nonminimal rules}
\cite{bradix99}: 
\begin{itemize}
\item 
\emph{If a rule $r'$ is an implication
of another rule $r$, then $r'$ can be removed from the original program.}
\end{itemize}

According to \cite{bradix99}, a rule $r$ is an implication of another rule
$r'$ if $\head{r'}\subseteq \head{r}$,
$\body{r'}\subseteq \body{r}$ and at least one inclusion is proper.

This program transformation seems quite intuitive at first glance. For example,
if we have a disjunctive program $P_{\ref{exa:travel:modi}}$ as follows
\begin{equation}\label{exa:travel:modi}
\begin{array}{crcl}
r_1: & b \vee l & \la & \naf p \\
r_3: & l & \la & \naf p
\end{array}
\end{equation}
Then $r_1$ is an implication of $r_3$ and thus
it is intuitive to remove the first rule from $P_{\ref{exa:travel:modi}}$.

However, the notion of implication is too weak as shown in Example~\ref{exa:travel}.
In fact, $r_2$ is stronger than $r_3$ but $r_1$ is not an implication of $r_2$
according to Brass and Dix's definition.

Therefore, it is necessary to strengthen the notion of implication so
that the application domains as in Example~\ref{exa:travel} can be correctly
handled.
That is, we want that $r_1$ is also an ``implication'' of $r_2$ while
$r_1$ is an implication of $r_3$.


This observation leads to the following strengthening of implication.

\begin{definition}\label{def:S:implication:eq}
$r'$ is an s-implication of $r$ if $r'\neq r$ and at least one of the following
two conditions is satisfied:
\begin{enumerate}
\item $r'$ is an implication of $r$:
$\head{r'}\subseteq \head{r}$,
$\body{r'}\subseteq \body{r}$ and at least one inclusion is proper; or
\item $r$ can be obtained by changing some negative body literals of $r'$
into head atoms and removing some body literals from $r'$ if necessary.
\end{enumerate}
\end{definition}
%
For instance, according to the second condition in 
Definition~\ref{def:S:implication:eq}, the rule
$b\vee l\la \naf p$ is an s-implication of the rule $l\vee p\la$
although $b\vee l\la \naf p$ is not an implication of $l\vee p\la$.
It should be pointed out that the notion
of s-implications does not mean we transform a disjunctive rule
with default negation into a positive rule.
Now we prepare to introduce our new transformation-based semantics.
According to~\cite{bradix99}, an abstract semantics can be defined
as follows.
%
\begin{definition} \label{def:semantics}
A \BD-semantics ${\cal S}$ is a mapping which assigns to every disjunctive
program $P$ a set ${\cal S}(P)$ of pure disjunctions
such that the following conditions are satisfied:
\begin{enumerate}
\item 
if $Q'$ is a sub-disjunction of pure disjunction $Q$ and
$Q'\in {\cal S}(P)$, then $Q\in {\cal S}(P)$;
\item 
if the rule $\texta\leftarrow$ is in $P$ for a (positive) disjunction $\texta$,
then $\texta\in {\cal S}(P)$;
\item
if $a$ is an atom and $a\not\in \head{P}$ (i.~e. $a$ does not 
appear in the rule heads of $P$), 
then $\naf a\in {\cal S}(P)$.  
\end{enumerate}
\end{definition}
%
In general, a semantics satisfying the above conditions
is not necessarily a {\em suitable} one because Definition
\ref{def:semantics} is still very general.

Moreover, as we argued above, it is meaningful to extend 
the set ${\bf T}_{\wfs}$ of program transformations defined in~\cite{bradix99}.
We accomplish this by introducing a new program transformation
called {\em Elimination of s-implications},
which extends Brass and Dix's Elimination of nonminimal rules.

The new set ${\bf T}^*_{\wfs}$ of program transformations is obtained by replacing
the Elimination of nonminimal rules in ${\bf T}_{\wfs}$ by
the Elimination of s-implications
(In the sequel, $P_1$ and $P_2$ are disjunctive programs): 
\begin{itemize}
\item {\bf Unfolding}:
$P_2$ is obtained from $P_1$ by unfolding
if there is a rule $\texta\la b, \textb,\naf \textc$ in $P_1$ such that
\begin{eqnarray*}
P_2 &=& P_1 -\{\texta\la b, \textb,\naf \textc\} \\
    & & \cup \{\texta\vee (\texta'-\{b\})\la \textb,\textb',
          \naf \textc,\naf \textc')\D \\
    & &\hskip 5mm \text{there is a rule of } P_1: \texta'
          \la \textb',\naf \textc'\text{ such that } b\in \texta'\}.
\end{eqnarray*}
\item {\bf Elimination of tautologies}:
 $P_2$ is obtained from  $P_1$ by
elimination of tautologies if there is a rule
$\texta\la \textb,\naf \textc$ in $P_1$ such that 
$\texta\cap \textb\neq\emptyset$
and $P_2=P_1-\{\texta\la \textb,\naf \textc\}$.

\item {\bf Elimination of s-implications}:
 $P_2$ is obtained from  $P_1$ by
elimination of s-implications if there are two distinct
rules $r$ and $r'$ of $P_1$ such that $r'$ is an s-implication of $r$
and $P_2=P_1-\{r'\}$.
\item {\bf Positive reduction}:
  $P_2$ is obtained from  $P_1$ by
positive reduction if there is a rule
$\texta\la \textb,\naf \textc$ in $P_1$ and $c\in \textc$ 
such that $c\not\in head(P_1)$ and
$P_2=P_1-\{\texta\la \textb,\naf \textc\}\cup 
\{\texta\la \textb,\naf (\textc-\{c\})\}$.
\item {\bf Negative reduction}:
  $P_2$ is obtained from  $P_1$ by
negative reduction if there are two rules
$\texta\la \textb,\naf \textc$ and $\texta'\la$ in $P_1$ 
such that $\texta'\subseteq \textc$
and $P_2=P_1-\{\texta\la \textb,\naf \textc\}$.
\end{itemize}
%
%
\begin{example}
Consider the disjunctive program $P_{\ref{eqn:travel}}$ in Example~\ref{exa:travel}.
Since $r_1$ is an s-implication of $r_2$, $P$ can be transformed
into the following disjunctive program $P'$ by the Elimination of s-implications:
\[
\begin{array}{rcl}
l \vee p & \la &
\end{array}
\]
Suppose that ${\mathcal S}$ is a \BD-semantics.
Then by Definition~\ref{def:semantics},
$l\vee p\in {\mathcal S}$ and $\naf b\in {\mathcal S}$.
\end{example}
%
Let us consider another example.
\begin{example}
$P$ consists of the following five rules:
\[
\begin{array}{crcl}
r_1: & p\vee p_1\vee p_2 & \la & \\
r_2: & p_1\vee p_2       & \la & q\\
r_3: & p_3 & \la & p, q, \naf p_4\\
r_4: & p_3\vee p_4 & \la &  \\
r_5: & w\vee q   & \la & w, \naf p \\
r_6: & q         & \la &
\end{array}
\]
Then we have a transformation sequence:
\begin{enumerate}
\item 
By the Unfolding, we can remove $q$ from $r_2$ and $r_3$, and obtain the following $P_1$:
\[
\begin{array}{crcl}
r_1: & p\vee p_1\vee p_2 & \la & \\
r_2: & p_1\vee p_2       & \la & \\
r_3: & p_3         & \la & p,\naf p_4\\
r_4: & p_3\vee p_4       & \la & \\
r_5: & w\vee q           & \la & w, \naf p \\
r_6: & q                 & \la &
\end{array}
\]
\item
By the Elimination of tautologies, we can remove $r_5$ and
obtain $P_2$:
\[
\begin{array}{crcl}
r_1: & p\vee p_1\vee p_2 & \la & \\
r_2: & p_1\vee p_2       & \la & \\
r_3: & p_3 & \la & p, \naf p_4\\
r_4: & p_3\vee p_4& \la & \\
r_6: & q         & \la &
\end{array}
\]
\item
By the Elimination of s-implications, we remove $r_1$ and
obtain $P_3$ (since $r_1$ is an s-implication of $r_2$):
\[
\begin{array}{crcl}
r_2: & p_1\vee p_2       & \la & \\
r_3: & p_3 & \la & p, \naf p_4\\
r_4: & p_3\vee p_4       & \la & \\
r_6: & q         & \la &
\end{array}
\]
\item
Again, by the Elimination of s-implications, we can remove $r_3$
and obtain $P_4$ (since $r_3$ is an s-implication of $r_4$):
\[
\begin{array}{crcl}
r_2: & p_1\vee p_2 & \la & \\
r_4: & p_3\vee p_4 & \la & \\
r_6: & q           & \la &
\end{array}
\]
\item
By the Positive reduction, we obtain
\[
\begin{array}{crcl}
r_2: & p_1\vee p_2 & \la & \\
r_4: & p_3\vee p_4 & \la & \\
r_6: & q           & \la &
\end{array}
\]
\end{enumerate}
By Definition~\ref{def:semantics}, if ${\cal S}$ is a \BD-semantics,
then ${\cal S}(P)$ contains the set of pure literals
\(
\{p_1\vee p_2, p_3\vee p_4, q, \naf w\}
\).
\end{example}

We say a semantics ${\mathcal S}$ allows a program
transformation $T$ (or equivalently, ${\mathcal S}$ is invariant
under $T$) if ${\mathcal S}(P_1)={\mathcal S}(P_2)$
for any two disjunctive programs $P_1$ and $P_2$
with $P_2=T(P_1)$.

Let ${\mathcal S}$ and ${\mathcal S}'$ be two \BD-semantics.
${\mathcal S}$ is {\em weaker} than ${\mathcal S}'$ if
${\mathcal S}(P)\subseteq {\mathcal S}'(P)$ for any
disjunctive program $P$.

We present the main definition of this section as follows.
%
\begin{definition}($\dwfsnew$)\label{def:dwfs}
The semantics $\dwfsnew$ for disjunctive programs is defined as the
weakest \BD-semantics allowing all program transformations in
${\bf T}^*_{\wfs}$.
\end{definition}
This definition is not constructive and thus it can not be directly
used to compute the semantics $\dwfsnew$ and thus a bottom-up procedure
will be given in the next section. In the rest of this section,
we show some properties of $\dwfsnew$, some of which are generalizations
of the corresponding ones for $\dwfs$ given in~\cite{bradix99}.

We first prove the following two fundamental lemmas.
%
\begin{lemma}\label{lem:nonempty}
There is a \BD-semantics that allows all the program transformations
in ${\bf T}^*_{\wfs}$.
\end{lemma}
\begin{proof}{lem:nonempty}
We can justify that $\wfds$ is a \BD-semantics and allows ${\bf T}^*_{\wfs}$
(see Proposition~\ref{prop:invariance}).
\end{proof}
%
\begin{lemma}\label{lem:intersection}
Let ${\bf C}$ be a non-empty class of \BD-semantics for disjunctive programs.
Then
\begin{enumerate}
\item The intersection
\(
\bigcap_{{\cal S}\in {\bf C}} {\cal S}
\)
is still a \BD-semantics.
\item For any program transformation $T$,
if ${\cal S}$ allows $T$ for each
\(
{\cal S}\in {\bf C}
\),
then
\(
\bigcap_{{\cal S}\in {\bf C}} {\cal S}
\)
also allows $T$.
\end{enumerate}
\end{lemma}
The proof of this lemma is direct and thus we omit it here.

Therefore, we have the following result which shows
that semantics $\dwfsnew$ assigns the unique model state
${\dwfsnew}(P)$ for each disjunctive program $P$.
\begin{theorem}\label{thm:dwfs:existence}
For any disjunctive program $P$, ${\dwfsnew}(P)$ is well-defined.
\end{theorem}
%
\begin{proof}{thm:dwfs:existence}
Let $\Sem({\bf T}^*_{\wfs})$ be the class of \BD-semantics that allow
${\bf T}^*_{\wfs}$. Then, by Lemma~\ref{lem:nonempty}, $\Sem({\bf T}^*_{\wfs})$
is non-empty. Furthermore, by Lemma~\ref{lem:intersection}, 
we have that 
\[
{\dwfsnew}(P)=\bigcap_{{\cal S}\in \Sem({\bf T}^*_{\wfs})} {\cal S}(P)
\]
\end{proof}
Since the set ${\bf T}_{\wfs}$ of program transformations
in~\cite{bradix99} is not stronger than ${\bf T}^*_{\wfs}$,
our $\dwfsnew$ extends the original D-WFS in the following sense.
%
\begin{theorem}\label{thm:dwfs:relation}
Let $P$ be a disjunctive program. Then
\[
\dwfs(P)\subseteq \dwfsnew(P).
\]
\end{theorem}
%
The converse of Theorem~\ref{thm:dwfs:relation} is not true
in general. As we will see in Section~\ref{section:coindence}, 
for the disjunctive program
in Example~\ref{exa:travel}, $\naf b\in \dwfsnew(P)$ but
$\naf b\not\in \dwfs(P)$.
This theorem also implies that $\dwfsnew$ extends the restricted STATIC
since the $\dwfs$ is equivalent to the restricted STATIC~\cite{brdinipr01}.

\section{Bottom-up Computation}\label{section:coindence}
\label{computation}
As shown in~\cite{bradix99}, the transformation-based approach
naturally leads to a bottom-up computation for the well-founded
semantics.
In this section, we will first provide a bottom-up procedure for $\dwfsnew$
and then show the equivalence of $\dwfsnew$ and $\wfdsnew$.
As a consequence, we also provide a bottom-up computation for $\wfdsnew$.

Let $P$ be a disjunctive program.
Our bottom-up computation for $\dwfsnew(P)$ consists of two stages.
At the first stage,  $P$ is equivalently transformed
into a negative program $\Lft(P)$ called the 
\emph{least fixpoint transformation} of $P$~\cite{bradix99,wan00}. 
The basic idea is to first evaluate body atoms of the rules in $P$ but delay
the negative body literals. The second stage is to further
reduce $\Lft(P)$ into another negative disjunctive $\res^*(P)$
from which the semantics
$\dwfsnew(P)$ can be directly read off.
\subsection{The Least Fixpoint Transformation}
In this subsection, we briefly recall the least fixpoint transformation.
The details of this notion can be found in~\cite{bradix99,wan00}.

We define the \emph{generalized disjunctive base} ${\GDB}_P$ of a
disjunctive logic program $P$ as the set of all conditional facts
whose atoms appear in $P$:
\[
\begin{array}{cl}
{\GDB}_P= &\{a_1\vee \cdots \vee a_r\leftarrow 
           \naf b_1,\ldots ,\naf b_s:\; a_i, b_j\in B_P, \\
          &i=1,\ldots ,r;j=1,\ldots ,s \text{ and } r>0, s\ge 0\}
\end{array}
\]


Having the notion of the generalized disjunctive base,
we are ready to introduce the immediate consequence operator $T_P^G$ 
for general disjunctive program $P$, which generalizes the immediate 
consequence operator for positive program $P$
(see Definition~\ref{def:imediate:operator:original}). 
The definition of the least fixpoint transformation will be based on this operator.

\begin{definition} \label{def:imediate:operator}
For any disjunctive program $P$, the generalized consequence operator
$T_P^G:2^{{\GDB}_P}\rightarrow 2^{{\GDB}_P}$ is defined as, for any $J\subseteq
{\GDB}_P$,
\[
T_P^G(J)
\;=\;
\left\{
  C\in {\GDB}_P\left|\; 
    \begin{array}{rl}
           & \text{there exist a rule}\;
       {\alpha}'\leftarrow b_1,\ldots,b_m,\naf b_{m+1},\ldots,\naf b_s
            \\
           & \text{in $P$ and } C_1, \ldots , C_m\in {\GDB}_P\text{ such that}
            \\
      (1). & b_i\vee head(C_i)\leftarrow body(C_i)
             \text{ is in } J, \text{ for all }i=1, \ldots ,m;
             \\
      (2). & C=\alpha'\vee head(C_1)\vee\cdots \vee head(C_m)
             \leftarrow body(C_1), \ldots , 
             \\
           & body(C_m), \naf b_{m+1},\ldots , \naf b_s
    \end{array}
  \right\}\right.
\]
\end{definition}

This definition looks a little tedious at first sight. In fact, its intuition is 
quite simple - it defines the following form of resolution:
$$
\frac{\alpha'\leftarrow b_1,\ldots ,b_m,\beta_1,\ldots ,\beta_s;\; 
      b_1\vee\alpha_1\leftarrow \beta_{11},\ldots ,\beta_{1t_1};\;
              \cdots ;
     b_m\vee\alpha_m\leftarrow \beta_{m1},\ldots ,\beta_{mt_m}
     }
    {\alpha'\vee\alpha_1\vee\cdots \vee\alpha_m\leftarrow \beta_{11},\ldots ,\beta_{1t_1},
\cdots , \beta_{m1}, \ldots ,\beta_{mt_m}, \beta_1,\ldots ,\beta_s
     }
$$
where $\alpha'$ and $\alpha$s with subscripts are positive disjunctive 
literals;  $\beta$s with subscripts are negative literals.

For any disjunctive program $P$, its generalized consequence operator
$T_P^G$ is continuous and hence possesses the least fixpoint 
$\Lft(P)=T_P^G\uparrow \omega$. Notice that $\Lft(P)$ is a negative disjunctive
program and is said to be the \emph{least fixpoint transformation} of $P$. 

For instance, consider the following disjunctive program $P$:
\begin{equation}
  \label{exa:travel:1}
  \begin{array}[t]{crcl}
b \vee l & \la & u,\naf p \\
l      & \la & v   \\
p \vee v & \la & u, \naf w  \\
u        & \la &  
   \end{array}
\end{equation}
Then its least fixpoint transformation $\Lft(P)$ is as follows:
\begin{equation}
  \label{exa:travel:2}
  \begin{array}[t]{crcl}
b \vee l & \la & \naf p \\
l \vee p     & \la & \naf w \\
p \vee v & \la & \naf w \\
u        & \la &  
   \end{array}
\end{equation}

Lemma 5.1 in~\cite{wan00} can be restated as the following.

\begin{lemma} \label{L501}
Let $\Delta$ be a hypothesis of disjunctive program $P$ 
(i.~e. $\Delta\subseteq \DB_P^-$) and $\alpha\in {\DB}_P^+$.
Then
\[
\Delta\vdash_P\alpha\;\text{if and only if}\; \Delta\vdash_{\Lft(P)}\alpha .
\]
\end{lemma}
By Lemma~\ref{L501}, it follows that the least fixpoint transformation $\Lft$
is invariant under the semantics $\wfdsnew$.
\begin{theorem}\label{thm:wfds:lft}
Let $\Lft(P)$ be the least fixpoint transformation of disjunctive program
$P$. Then
\[
\wfdsnew(Lft(P))=\wfdsnew(P).
\] 
\end{theorem}
It has been proven in~\cite{bradix99} that $\Lft$ also is invariant under the 
transformation-based semantics.

\begin{theorem} \label{thm:preserve:trans}
If a \BD-semantics ${\mathcal S}$ allows the Unfolding and Elimination of tautologies,
then ${\mathcal S}(\Lft(P))={\mathcal S}(P)$.
\end{theorem}
By Theorem~\ref{thm:preserve:trans}, it is direct that the least fixpoint 
transformation $\Lft$ is invariant under the semantics $\dwfsnew$.
\begin{corollary}\label{thm:Lft:preserve:dwfsnew}
Let $\Lft(P)$ be the least fixpoint transformation of disjunctive program
$P$. Then
\[
\dwfsnew(Lft(P))=\dwfsnew(P).
\] 
\end{corollary}
%
\subsection{Strong Residual Program}
In general, the negative program $\Lft(P)$ can be further
simplified by deleting unnecessary rules. This leads to the idea of so-called
\emph{reductions}, which was firstly studied in~\cite{bry90} and
then generalized to the case of disjunctive logic programs 
in~\cite{bradix99}. The logic program obtained by fully performing reduction
on a disjunctive program $P$ is called the residual program of $P$. 

In the following we define the notion of \emph{strong residual programs},
which is a generalization of Brass and Dix's residual programs.

The {strong reduction operator} $R^*$ is defined as,
for any negative program $N$ (i.~e. a set of conditional facts),
\[
R^*(N)
\;=\;
\left\{
  \texta\la \naf (\textc\cap \head{N})\left|\; 
    \begin{array}{rl}
           & \text{ there is a rule $r\in N$}: \texta\la \naf \textc 
          \text{ such that}
            \\
           & \text{$r$ is not an s-implication of $r'\neq r$ for any rule
             $r'\in N$}
    \end{array}
  \right\}\right.
\]
The intuition behind the above operator is very simple: We first select all
minimal rules wrt s-implication and then remove all negative body literals
whose atom does not appear in a rule head.
Since an implication under Brass and Dix's sense is also an s-implication,
we have that $R^*(N)\subseteq R(N)$ for any negative program $N$.
The strong reduction $R^*$ is really different from Brass and Dix's reduction $R$.
For example, the reduction of program $P_{\ref{eqn:travel}}$ is itself while
its strong reduction is $l\vee p\leftarrow$.
The reduction operator $R^*$ also possesses
a more elegant form than the one defined in~\cite{bradix99}.
 
For any disjunctive program $P$, we can first transform
it into the negative disjunctive program $\Lft(P)$. 
Then, \emph{fully} perform the reduction $R^*$ on $\Lft(P)$ to obtain 
a simplified negative program $res^*(P)$ (the \emph{strong residual program} of $P$). 
The iteration procedure of $R^*$ will finally stop in finite steps
because $B_P$ contains finite number of atoms and the total number of
atoms occurring in each $N$ is reduced by $R^*$.
This procedure is precisely formulated in the next definition, which is in
a similar form as Definition 3.4 in~\cite{bradix99} (the difference
is in that we have a new reduction operator $R^*$ here).
%
\begin{definition}(strong residual program)\label{def:strong:res}
Let $P$ be a disjunctive program. Then we have a sequence of negative
programs $\{N_i\}_{i\ge 0}$ with $N_0=\Lft(P)$ and $N_{i+1}=R^*(N_i)$.
If $N_t$ is a fixpoint of $R^*$, i.~e. $N_t=R^*(N_t)$,
then we say $N_t$ is the strong residual program
of $P$ and denote it as $\res^*(P)$. 
\end{definition}

We first show that the strong residual program is invariant under all the
program transformations in ${\bf T}^*_{\wfs}$. 
This result actually means that the problem of evaluating $P$ can be reduced to 
that of evaluating its strong residual program.
\begin{theorem}\label{thm:res:equal}
Let $P$ and $P'$ be two disjunctive programs. If
$P$ is transformed into $P'$ by a program transformation $T$
in ${\bf T}^*_{\wfs}$, then
$res^*(P)=res^*(P')$.
\end{theorem}
To prove this theorem, we need the following lemma,
which is a reformulation of Lemma 4.1 and 4.2 in~\cite{bradix99}.
\begin{lemma}\label{lem:bradix:4}
If $P$ is transformed into $P'$ by either the Unfolding or the Elimination of Tautology,
then 
\begin{enumerate}
  \item 
    $P$ and $P'$ have the same set of minimal models;
  \item
    $\Lft(P)$ and $\Lft(P')$ contain the same set of minimal conditional facts.
  \end{enumerate}
\end{lemma}
Here, we say a conditional fact $A\leftarrow\naf C$ is minimal in a set $N$ of
conditional facts if there is no conditional fact $A'\leftarrow\naf C'$ in $N$
such that $A'\subseteq A$, $C'\subseteq C$ and at least one inclusion is strict.
\begin{proof}{thm:res:equal}
\begin{enumerate}
\item If $T$ is either the Unfolding or the Elimination of Tautology,
then by Lemma~\ref{lem:bradix:4}(1), $P$ and $P'$ have the same set of minimal models.
Thus, by Lemma~\ref{lem:bradix:4}(2), 
$\Lft(P)$ and $\Lft(P')$ contain the same set of minimal conditional facts.
Thus, $res^*(P)=res^*(P')$.

\item If $T$ is the Elimination of s-implications, then there are two rules $r$ and $r'$ 
in $P$ such that $r'$ is an s-implication of $r$ and $P'=P-\{r'\}$.

We show that if $r'_1$ is a resolvent of $r'$ with another rule $r''$
by the resolution rule (\ref{rule:sli}),
then $r'_1$ is an s-implication of $r_1$ where $r_1$ is either $r$ or 
a resolvent of $r$ with $r''$ by (\ref{rule:sli}).
 
  To prove this, we need only to consider the following cases:
  \begin{enumerate}
  \item Let $r'$ is of form 
    \(
    A'\vee a\la\body{r'}
    \),
    $r''$ is of form
    \(
    \head{r''}\la a,B,\nbody{r''}
    \)
    and
    $r'_1$ is obtained by resolving the head atom $a$ of $r'$ with the body atom $a$ of $r''$.
    That is, $r'_1$ is of form
    \(
    A'\vee \head{r''}\la \body{r'},B,\nbody{r''}
    \):

    If $a$ does not appear in $\head{r}$, then it is obvious that $r'_1$ is s-implication of $r$;
    otherwise, we assume that $\head{r}$ is of form $A\vee a$ such that $A\subseteq A'$.
    
    Notice that the resolvent $r_1$ of $r''$ with $r$ on $a$ is of form
    \[
    A\vee \head{r''}\la \body{r},B,\nbody{r''}.
    \]
    Therefore, $r'_1$ an s-implication of $r_1$.
  \item Let $r'$ is of form 
    \(
     \head{r'}\la b, B',\nbody{r'}
    \),
    $r''$ is of form
    \(
    A''\vee b\la \body{r''}
    \)
    and
    $r'_1$ is obtained by resolving the body atom $b$ of $r'$ with the head atom $b$ of $r''$.
    That is, $r'_1$ is of form
    \(
    \head{r'}\vee A''\la \nbody{r''}, B',\nbody{r'}
    \).
    
    If $b$ does not appear in the body of $r$, then it is obvious that $r'_1$ is an 
    s-implication of $r$; otherwise, assume that $b\in\pbody{r}$.

    Notice that the resolvent $r_1$ of $r$ with $r''$ on $b$ is of form
    \[
    \head{r}\vee A''\la \nbody{r''}, B,\nbody{r}.
    \]
    Therefore, $r'_1$ an s-implication of $r_1$.
  \end{enumerate}

 Thus, we have that $res^*(P)=res^*(P')$.

\item If $T$ is the Positive reduction, then there is a rule
$\texta\la \textb,\naf \textc$ in $P_1$ and $c\in \textc$ 
such that $c\not\in head(P_1)$ and
$P_2=P_1-\{\texta\la \textb,\naf \textc\}\cup 
\{\texta\la \textb,\naf (\textc-\{c\})\}$.

By the definition of the least fixpoint transformation, the only difference of $\Lft(P)$
from $\Lft(P')$ is in the rules whose body contain $\naf c$.
Moreover, for each conditional fact $r$ in $\Lft(P)$ of form $A\leftarrow\naf C,\naf c$, 
there is the corresponding conditional fact $A\leftarrow\naf C$ in $\Lft(P')$.
Thus, by definition of the strong reduction, we have
that $\Lft(P)=\Lft(P')$.
\end{enumerate}
\end{proof}
This theorem has the following interesting corollary.
\begin{corollary}\label{thm:res:allow}
Let ${\mathcal S}$ be a \BD-semantics allowing 
${\mathcal S}(P)={\mathcal S(\res^*(P)})$ for all disjunctive program $P$.
Then ${\mathcal S}$ allows all program transformations in ${\bf T}^*_{\wfs}$.
\end{corollary}
\begin{proof}{thm:res:allow}
Suppose that a disjunctive program $P$ is transformed into another disjunctive
program $P'$. By Theorem~\ref{thm:res:equal}, we have $res^*(P)=res^*(P')$.
Thus, 
\[
{\mathcal S}(P)={\mathcal S}(res^*(P))={\mathcal S}(res^*(P'))
={\mathcal S}(P').
\]
\end{proof}
This corollary implies that, if ${\mathcal S}_0$ is a mapping 
from the set of all strong residual programs to the set of model states
and it satisfies all properties in Definition~\ref{def:semantics},
then the mapping defined by ${\mathcal S}(P)={\mathcal S(\res^*(P)})$
is a \BD-semantics. 

Before we show the main theorem of this section, we need two lemmas.
\begin{lemma}\label{lem:residual:sem}
Given disjunctive program $P$, we have 
\[
{\dwfsnew}(\res^*(P))=S_0^+(P)\cup S_0^-(P)
\]
where

\noindent
\(
S_0^+(P)=\{\texta\in {\DB}_P^+\;\D \text{rule $\texta'\leftarrow$ 
                 is in $res^*(P)$ for some  sub-disjunction $\texta'$ of $\texta$}\}
\)

\noindent
\(
S_0^-(P)=\{\texta\in {\DB}_P^- \D \text{if $a\not\in \head{res^*(P)}$ 
                for some atom $a$ appearing in $\texta$.}\}
\)
\end{lemma}
Thus, for any disjunctive program $P$, it is an easy task to compute the
semantics $\dwfsnew(\res^*(P))$ of its strong residual program. 
\begin{proof}{lem:residual:sem}
Define a mapping ${\mathcal S}_0$ from disjunctive programs to model states
as follows:
\[
{\mathcal S}_0(P)=S_0^+(P)\cup S_0^-(P).
\]
Then ${\mathcal S}_0$ is a \BD-semantics by Definition~\ref{def:semantics}
and ${\mathcal S}_0$ allows ${\bf T}^*_{\wfs}$ by Corollary~\ref{thm:res:allow}.

Since ${\dwfsnew}(\res^*(P))$ is the least \BD-semantics that allows ${\bf T}^*_{\wfs}$,
we have that
\[
{\dwfsnew}(\res^*(P))\subseteq {\mathcal S}_0(\res^*(P)).
\]

On the other hand, by Definition~\ref{def:semantics},
\[
{\mathcal S}_0(\res^*(P))\subseteq {\dwfsnew}(\res^*(P)).
\]
Therefore, the conclusion of the lemma is correct.
\end{proof}
The next lemma says that $P$ is equivalent to $res^*(P)$ under the semantics $\dwfsnew$.
\begin{lemma}\label{lem:residual:eqv}
For any disjunctive program $P$, we have
\[
{\dwfsnew}(P)={\dwfsnew}(\res^*(P)).
\]
\end{lemma}
\begin{proof}{lem:residual:eqv}
By Corollary~\ref{thm:Lft:preserve:dwfsnew}, it suffices to prove that the conclusion 
holds for all negative disjunctive programs.
Let $N$ be an arbitrary negative program. 
We want to show that $N$ is equivalent to $R^*(N)$ under the semantics $\dwfsnew$.
That is, $\dwfsnew(N)=\dwfsnew(R^*(N))$. This can be further reduced to show
that $N$ can be transformed into $R^*(N)$ by the transformations in ${\bf T}^*_{\wfs}$.

Notice that $R^*(N)$ is obtained from $N$ by removing some rules and/or remove
some body atoms. There are two possibilities by the definition of $R^*$:
\begin{enumerate}
\item A rule $r\in N$ is removed due to that $r$ is an s-implication of $r'$ for some $r'\in N$:
This removal can be directly simulated by the Elimination of s-implications;
\item A negative literal $\naf a$ is removed from the body of a rule $r\in N$
due to
$a\not\in\head{N}$: This removal can be simulated by the Positive reduction.
\end{enumerate}
Thus, $N$ can be transformed into $R^*(N)$ through the transformations in ${\bf T}^*_{\wfs}$.
\end{proof}
The main theorem in this section can be stated as follows,
which tell us that the evaluation of $P$ under $\wfdsnew$ can be reduced to that
of the strong residual program (the latter is an easy job as we have seen).
\begin{theorem}\label{thm:comp:dwfsnew}
For any disjunctive program $P$, we have
\[
{\dwfsnew}(P)=S_0^+(P)\cup S_0^-(P)
\]
where

\noindent
\(
S_0^+(P)=\{\texta\in {\DB}_P^+\;\D \text{rule $\texta'\leftarrow$ 
                 is in $res^*(P)$ for some  sub-disjunction $\texta'$ of $\texta$}\}
\)

\noindent
\(
S_0^-(P)=\{\texta\in {\DB}_P^- \D \text{ if $a\not\in \head{res^*(P)}$ 
                for some atom $a$ appearing in $\texta$.}\}
\)
\end{theorem}
\begin{proof}{thm:comp:dwfsnew}
It follows directly from Lemma~\ref{lem:residual:sem} and~\ref{lem:residual:eqv}.
\end{proof}
\begin{example}
Consider again the disjunctive program $P_{\ref{eqn:travel}}$ in Example~\ref{exa:travel}.
The strong residual program $res^*(P)$ is as follows:
\[
\begin{array}{rcl}
l & \la & \naf p \\
l \vee p & \la &
\end{array}
\]
Thus,
$\dwfsnew(P)=\{l \vee p, \naf b\}$
\footnote{$\dwfsnew (P)$ should include all pure disjunctions
implied by either $l \vee p $ or $\naf b$. However, the
little abusing of notion here simplifies our notation.}.
\end{example}
%
\subsection{Equivalence of $\wfdsnew$ and $\dwfsnew$}
Before we present the main theorem of this section,
we need some properties of $\wfdsnew$.
First, we can justify that $\wfdsnew$ is a semantics in the sense of 
Definition~\ref{def:semantics}. Moreover, it possesses
the following two important properties.
%
\begin{proposition} \label{prop:invariance}
$\wfdsnew$ allows all program transformations in ${\bf T}^*_{\wfs}$.
\end{proposition}
This proposition implies that the argumentation-based semantics
$\wfdsnew$ is always at least as strong as 
the transformation-based semantics $\dwfsnew$.

\begin{proof}{prop:invariance}
  \begin{enumerate}
  \item If $P_1$ is transformed into $P_2$ by either the Unfolding or
    Elimination of tautologies, then
    $\Lft(P_1)=\Lft(P_2)$.
   
  Therefore, $\wfds(P_1)=\wfds(\Lft(P_1))=\wfds(\Lft(P_2))=\wfds(P_2)$.

  \item If $P_2$ is obtained from  $P_1$ by
   Positive reduction, then there is a rule
  $\texta\la \textb,\naf \textc$ in $P_1$ and $c\in \textc$ 
  such that $c\not\in head(P_1)$ and
  $P_2=P_1-\{\texta\la \textb,\naf \textc\}\cup 
  \{\texta\la \textb,\naf (\textc-\{c\})\}$.

  We need only to show that $\wfds(P_1)$ and $\wfds(P_2)$ contain the same set
  of negative literals. That is, $\wfdh(P_1)=\wfdh(P_2)$.
  We use induction on $k$ to show 
  \begin{equation}\label{eqn:pos:reduc}
    {\mathcal A}^k_{P_1}(\emptyset)={\mathcal A}^k_{P_2}(\emptyset)
  \end{equation}
  for any $k\ge 0$.

  For $k=0$, it is obvious.

  Assume that (\ref{eqn:pos:reduc}) holds for $k$, we want to prove
  (\ref{eqn:pos:reduc}) also holds for $k+1$.

  Notice that $\naf c$ belongs to both $\wfds(P_1)$ and $\wfds(P_2)$.
  In particular, $\naf c\in {\mathcal A}_{P_t}(\emptyset)$ for $t=1,2$.

  Let $\naf p\in {\mathcal A}^{k+1}_{P_1}(\emptyset)$.
  For any hypothesis $\Delta'$ with $\Delta'\leadsto_{P_2}\{\naf p\}$,
  we have that
  \[
  (\Delta'\cup\{\naf c\})\leadsto_{P_1}\{\naf p\}.
  \]
  Thus,
  \[
  {\mathcal A}^k_{P_1}(\emptyset)\leadsto_{P_1}(\Delta'\cup\{\naf c\}).
  \]
  By induction, 
  \[
  {\mathcal A}^k_{P_2}(\emptyset)\leadsto_{P_1}(\Delta'\cup\{\naf c\}).
  \]
  Since $\naf c\in {\mathcal A}^k_{P_2}(\emptyset)$,
  we have  ${\mathcal A}^k_{P_2}(\emptyset)\leadsto_{P_2}\Delta'$.
  
  Therefore, $\naf p\in {\mathcal A}^{k+1}_{P_2}(\emptyset)$.

  This implies
  \[
  {\mathcal A}^{k+1}_{P_1}(\emptyset)\subseteq{\mathcal A}^{k+1}_{P_2}(\emptyset)
  \]

  For the converse inclusion, let $\naf p\in {\mathcal A}^{k+1}_{P_2}(\emptyset)$.
  For any hypothesis $\Delta'$ with $\Delta'\leadsto_{P_1}\{\naf p\}$,
  then $(\Delta'\cup\{\naf c\})\leadsto_{P_2}\{\naf p\}$.

  This means ${\mathcal A}^k_{P_2}(\emptyset)\leadsto_{P_1}(\Delta'\cup\{\naf c\})$.
  
  Since $\naf c\in {\mathcal A}^k_{P_2}(\emptyset)$, we have
  ${\mathcal A}^k_{P_2}(\emptyset)\leadsto_{P_1}\Delta'$.

  That is, $\naf p\in {\mathcal A}^{k+1}_{P_1}(\emptyset)$.
  
Thus,
  \[
  {\mathcal A}^{k+1}_{P_2}(\emptyset)\subseteq{\mathcal A}^k_{P_1}(\emptyset)
  \]
  \item If $P_2$ is obtained from  $P_1$ by the Elimination of s-implications,
    then there are two possible subcases:
    \begin{enumerate}
    \item There are two rules $r_1\in P_1$ of form
     \(
     A_1\vee A_2\leftarrow B,\naf C,\naf c_1,\ldots,\naf c_t
     \)
     and $r_2\in P_1$ of form
     \(
     A_1\vee c_1\vee\cdots\vee c_t\leftarrow B',\naf C'
     \)
   such that $B'\subseteq B$, $C'\subseteq C$ and $P_2=P_1-\{r_1\}$:

It suffices to show that for any two hypotheses $\Delta$ and $\Delta'$,

$\Delta\leadsto_{P_1}\Delta'$
iff
$\Delta\leadsto_{P_2}\Delta'$.

This can be reduced to show that, for any hypothesis $\Delta$ and any positive disjunction $A$,

$\Delta\vdash_{P_1} A$
iff
$\Delta\vdash_{P_2} A$.

It is direct that $\Delta\vdash_{P_2} A$
implies
$\Delta\vdash_{P_1} A$
since $P_2\subseteq P_1$.

Let $\Delta\vdash_{P_1} A$.

If $r_1$ is not involved in the derivation of $\Delta\vdash_{P_1} A$, it is trivial that
$\Delta\vdash_{P_2} A$. 

If $r_1$ is involved in the derivation of $\Delta\vdash_{P_1} A$, then $r_1$ must be
revolved into a rule $r_3$ of form
\(
A_1\vee A_2\vee A_3\leftarrow\naf C,\naf C',\naf c_1,\ldots,\naf c_t.
\)
That is, all body atoms should be resolved with other rules until there is no body atom. 

If a head atom of $r_2$ is resolved with a body literal of another rule $\bar{r}$,
then we get a rule of form 
\begin{equation}\label{eqn:1:rule}
A'_1\vee A'_2\vee\head{\bar{r}}\leftarrow \naf C'',\naf c_1,\ldots,\naf c_t.
\end{equation}
where $A'_1\subseteq A_1$.

On the other hand, if we replace $r_1$ with $r_2$ in the above derivation,
we will get a rule of form
\begin{equation}\label{eqn:2:rule}
A'_1\vee\head{\bar{r}}\vee c_1\vee\cdots\vee c_t\leftarrow \naf C'''.
\end{equation}
such that $C'''\subset C''$.

If $\Delta$ derives $A$ through the rule (\ref{eqn:1:rule}),
then $\{\naf c_1,\ldots,\naf c_t\}\subseteq\Delta$.

Notice that $A'_1\vee\head{\bar{r}}$ is a sub-disjunction of $A'_1\vee A'_2\vee\head{\bar{r}}$,
thus $\Delta$ can also derives $A$ without $r'$ (through $r$).

This implies that $\Delta\vdash_{P_2} A$.
    \item There are two rules $r$ and $r'$ in $P_1$ such that
      $\head{r'}\subseteq\head{r}$, $\body{r'}\subseteq\body{r}$ and 
      $P_2=P_1-\{r'\}$: Similar to Case 1, we can prove that
      for any positive disjunction $A$ and any hypothesis $\Delta$,
   \(
   \Delta\vdash_{P_1} A
   \)
   iff
   \(
   \Delta\vdash_{P_2} A
   \).
  This implies that $\wfds(P_1)=\wfds(P_2)$.
    \end{enumerate}
\end{enumerate} 
\end{proof}
The next result convinces that the strong residual program 
$res^*(P)$ of disjunctive program $P$ is equivalent to
$P$ \wrt the semantics $\wfdsnew$. Therefore, we can first
transform $P$ into $res^*(P)$ and then compute $\wfdsnew (res^*(P))$.
%
\begin{proposition}\label{pro:residual}
For any disjunctive program $P$, 
\[
\wfdsnew (P)=\wfdsnew (res^*(P)).
\]
\end{proposition}
\begin{proof}{pro:residual}
By Lemma~\ref{thm:wfds:lft}, $\wfds(P)=\wfds(\Lft(P))$.

Similar to Lemma~\ref{lem:residual:eqv}, we know that $R^*$ can be simulated by 
${\bf T}^*_{\wfs}$. 

Thus, $\wfds(\Lft(P))=\wfds(\res^*(P))$.

That is, $\wfds(P)=\wfds(\res^*(P))$.
\end{proof}
Now we can state the main result of this section, which
asserts the equivalence of $\dwfsnew$ and $\wfdsnew$. 
%
\begin{theorem}\label{thm:equiv1}
For any disjunctive logic program $P$, 
\[
\wfdsnew (P)={\dwfsnew}(P).
\]
\end{theorem}
An important implication of this result is that the well-founded semantics 
$\wfdsnew$ also enjoys a bottom-up procedure similar to the $\dwfs$.
\begin{proof}{thm:equiv1}
For simplicity, we denote $res^*(P)$ by $N$ throughout
this proof. 

By Proposition~\ref{pro:residual}, it suffices to show that
$\wfdsnew (N)={\dwfsnew}(N)$ for any disjunctive program $P$.

First, from Proposition~\ref{prop:invariance}, it follows that
$\wfdsnew (N)\supseteq {\dwfsnew}(N)$.

We want to show the converse inclusion:
$\wfdsnew (N)\subseteq {\dwfsnew}(N)$.

Let ${\texta}\in \wfdsnew (N)$, we consider two cases:

Case 1. ${\texta}$ is a negative disjunction:
then $\naf a$ is in $\wfdsnew (N)$ for some atom $a$ in ${\texta}$.
It suffices to show that
$a\not\in head(N)$ for any negative literal
$\naf a$ in ${\mathcal A}_N^k$ for $k\ge 0$.

We use induction on $k$.

It is obvious for $k=0$.

Assume that the above proposition holds for
$k$, we want to show that it also holds for $k+1$.

Let $\naf a\in {\mathcal A}_N^{k+1}$.
On the contrary, suppose that
there is a rule $r:\; a\vee \texta'\leftarrow \naf \textc'$ in $N$. 
Denote $\Delta'=\{\naf p \D p\in atoms(\texta')\cup \textc'\}$,
then  $\Delta'\leadsto_{N} \{\naf a\}$.
Thus ${\mathcal A}_N^k\leadsto_N \Delta'$.
This means that ${\mathcal A}_N^k\vdash_N c_1\vee\cdots\vee c_t$
for some atoms $c_1,\ldots,c_t$ appearing in $\Delta'$ and $t>0$.
Therefore, there is a rule 
$c_1\vee\cdots\vee c_t\vee \texta''\leftarrow \naf \textc''$
such that $\Delta''\subseteq {\mathcal A}_N^k$
where $\Delta''=\{\naf p \D p\in atoms(\texta'')\cup \textc''\}$. 
By the induction assumption,
$\texta''=\emptyset$; by $N=res^*(P)$, $\textc''=\emptyset$.
This contradicts to the fact $N=res^*(P)$.
Therefore, $a\not\in head(N)$.

Case 2. ${\texta}$ is a positive disjunction:
then there is a rule in $N$: 
$\texta'\vee \texta''\leftarrow \naf \textc''$
such that $\texta' \subseteq \texta$ and 
$\Delta''\subseteq \wfdsnew(N)$ where $\Delta''$ has the same form as in Case 1.
Parallel to Case 1, we can prove that
$\texta''=\emptyset$ and since $N=res^*(P)$,
$\textc''=\emptyset$. This implies that the rule
$\texta'\leftarrow$ is in $N$.
\end{proof}
\section{Unfounded Sets} \label{section:unfounded}
The first definition of the well-founded model~\cite{varosc91}
is given in terms of \emph{unfounded sets} and it has been proven
that the notion of unfounded sets constitutes a powerful and intuitive tool 
of defining semantics for logic programs.
This notion has also been generalized to characterize stable semantics
for disjunctive logic programs in~\cite{eilesa97,lerusc97}.
However, the two kinds of unfounded sets defined 
in~\cite{eilesa97,lerusc97} can not be used to define an intended 
well-founded semantics for disjunctive
programs. 
%
\begin{example}\label{exa:dix}
\footnote{This example is due to J\"urgen Dix
(personal communication).}
\[
\begin{array}{rcl}
a\vee b & \la & \\
c       & \la & \naf a, \naf b
\end{array}
\]
\end{example}
%
Intuitively, $\naf c$ should be derived from the above disjunctive program
and actually, many semantics including DWFS, STATIC and WFDS assign
a truth value \lq {\em false}' for $c$.
However, according to the definitions of unfounded sets
in \cite{lerusc97,eilesa97}, $c$ is not in any $n$-fold application
of the well-founded operators on the empty set.
For this reason, a more reasonable definition of the unfounded sets
for disjunctive programs is in order.

In this section, we will define a new notion of unfounded sets
for disjunctive programs and show that 
the well-founded semantics U-WFS defined by 
our notion is equivalent to $\dwfsnew$ and $\wfdsnew$.

We say $\body{r}$ of $r\in P$ is {\em true} 
wrt model state S, denoted $S\models \body{r}$,
if $\body{r}\subseteq S$;
$\body{r}$ is {\em false} 
wrt model state $S$, denoted $S\models \neg \body{r}$
if either 
(1) the complement of a literal in $\body{r}$ is in $S$ or
(2) there is a disjunction $a_1\vee\cdots\vee a_n\in S$ such that
$\{\naf a_1,\ldots,\naf a_n\}\subseteq \body{r}$.

In Example~\ref{exa:dix}, the body of the second rule is false
wrt $S=\{a\vee b\}$.
%
\begin{definition}\label{def:unfounded}
Let $S$ be a model state of disjunctive program $P$,
a set $X$ of ground atoms is an {\em unfounded set}
for $P$ wrt $S$ if, for each $a\in X$ and each rule $r\in P$ 
such that $a\in \head{r}$, 
at least one of the following conditions holds:

1. the body of $r$ is false wrt $S$;

2. there is $x\in X$ such that $x\in body^+(r)$;

3. if $S\models \body{r}$, then
$S\models (\head{r}-X)$.
\end{definition}
%
Notice that the above definition generalized the notions
of unfounded sets in~\cite{eilesa97,lerusc97} in two ways.
Firstly, the original ones are defined only for interpretations
(sets of ground literals) rather than for model states.
An interpretation is a model state but not vice versa.
Secondly, though one can redefine the original notions of
unfounded sets for model states, such unfounded sets
are still too weak to capture the intended well-founded semantics of
some disjunctive programs. 
Consider Example~\ref{exa:dix}, let $S=\{a\vee b\}$.
According to Definition~\ref{def:unfounded}, 
the set $\{c\}$ is an unfounded set of $P$
wrt $S$, but $\{c\}$ is not an unfounded set in the sense of
Leone et al or Eiter et al.

Having the new notion of unfounded sets,
we are ready to define the well-known operator ${\mathcal W}_P$
for any disjunctive program $P$.

If $P$ has the greatest unfounded set wrt a model state $S$,
we denote it ${\mathcal U}_P(S)$. 
However, ${\mathcal U}_P(S)$ may be undefined for some $S$. For example,
let $P=\{a\vee b\}$ and $S=\{a,b\}$. Then $X_1=\{a\}$ and
$X_2=\{b\}$ are two unfounded sets wrt $S$ but $X=\{a,b\}$
is not. 
\begin{definition}
Let $P$ be a disjunctive program, the operator ${\mathcal T}_P$
is defined as, for any model state $S$,
\begin{eqnarray*}
{\mathcal T}_P(S) & = & \{\texta\in DB_P \;|\text{ there is a rule }
                         r\in P:\texta\vee a_1\vee\cdots\vee a_n \la \body{r} 
            \text{ such that } \\
       & &  S\models \body{r} \text{ and } \naf a_1,\ldots,\naf a_n\in S\}.
\end{eqnarray*}
\end{definition}
Notice that ${\mathcal T}_P(S)$ is a set of positive disjunctions
rather than just a set of atoms.
%
\begin{definition}
Let $P$ be a disjunctive program, the operator ${\mathcal W}_P$
is defined as, for any model state $S$,
\[
{\mathcal W}_P(S)={\mathcal T}_P(S)\cup \naf\!\!.{\mathcal U}_P(S).
\]
where $\naf\!\!.{\mathcal U}_P(S)=\{\naf p \D p\in {\mathcal U}_P(S)\}$.
\end{definition}
%
In general, ${\mathcal W}_P$ is a partial function
because there may be no greatest unfounded set wrt 
model state $S$ as mentioned previously.
 
However, we can prove that ${\mathcal W}_P$ has the least fixpoint.
Given a disjunctive program $P$, we define a sequence
of model states $\{W_k\}_{k\in {\mathcal N}}$
where $W_0=\emptyset$ and
$W_{k}={\mathcal W}_P(W_{k-1})$ for $k>0$.


Similar to Proposition 5.6 in \cite{lerusc97}, we can
prove the following proposition.
%
\begin{proposition}\label{pro:unfounded:exi}
Let $P$ be a disjunctive program. Then

1. Every  model state $W_k$
is well-defined and the sequence $\{W_k\}_{k\in {\mathcal N}}$
is increasing. 

2. the limit $\cup_{k\ge 0}W_k$ of the sequence 
$\{W_k\}_{k\in {\mathcal N}}$ is the least fixpoint of
${\mathcal W}_P$.
\end{proposition}
%
Before proving this proposition, we need the following lemmas.
%
\begin{lemma}\label{lem:unfounded:empty}

  \begin{enumerate}
  \item For each $k\ge 0$, $W_k$ is a consistent model state.
  \item For any $k\ge 0$ and any unfounded set $X$ for $P$ wrt $W_k$,
we have $W_k\cap X=\emptyset$.
  \end{enumerate} 
\end{lemma}
%
\begin{proof}{lem:unfounded:empty}
We prove this lemma by using simultaneous induction on $k$.
\paragraph{Base}\quad The conclusion is obviously true for $k=0$.

\paragraph{Induction} \quad
  \begin{enumerate}
  \item On the contrary, suppose that $W_k$ is not a consistent model state.
Then there is a positive disjunction
$a_1\vee\cdots\vee a_n\in W_k$ ($n\ge 1$) such that $\naf a_i\in W_k$ for any
$1\le i\le n$.

Since $a_1\vee\cdots\vee a_n\in W_k={\cal W}(W_{k-1})$,
there exists a rule $r\in P$:
\[
a_1\vee\cdots\vee a_n\vee A\la\body{r}
\]
such that $W_{k-1}\models\body{r}$ and
$\naf a\in W_{k-1}$ for any $a\in A$.

By the induction assumption, $W_{k-1}$ is a consistent model state,
we know that none of the conditions (1) and (2) in Definition~\ref{def:unfounded}
is satisfied by $X={\cal U}(W_{k-1})$ wrt $S=W_{k-1}$ and $r$.

Because $X$ is an unfounded set, the condition (3) in Definition~\ref{def:unfounded}
should be satisfied by $X={\cal U}(W_{k-1})$ wrt $S=W_{k-1}$ and $r$.
Thus, $W_{k-1}\models(\head{r}-X)$. 
Note that, since $a_i\in X$ for $i=1,\ldots,n$,
\(
\head{r}-X=A-X
\).

On the other hand, by induction assumption, $W_{k-1}$ is consistent
and thus, $W_{k-1}\not\models (A-X)$ since $\naf a\in W_{k-1}$ for any $a\in A$,
contradiction.

  \item On the contrary, suppose that there is an unfounded set $X$ for $P$ wrt $W_k$
such that $W_k\cap X\not=\emptyset$. 

Note that, by induction assumption, $W_{k-1}\cap X=\emptyset$.

Let $a\in W_k\cap X$. Then $a\in X$ implies the three conditions
in Definition~\ref{def:unfounded}
are satisfied by $r$ wrt $S=W_k$ and $X$. 

On the other hand, by assumption, $a\in W_k\setminus W_{k-1}$.
Since $a\in W_k={\cal T}(W_{k-1})$,
there exists a rule $r\in P$ such that
$W_{k-1}\models \body{r}$ and
$\naf b\in W_{k-1}$ for any $b\in(\head{r}-{a})$. 
This directly implies that
none of the conditions (1) and (2) in Definition~\ref{def:unfounded}
is satisfied by $r$ wrt $S=W_k$ and $X$. 
Since $W_{k-1}\subseteq W_k$, we have $W_k\models \body{r}$ and 
$\naf b\in W_k$ for any $b\in(\head{r}-{a}$. 

From the first part of this lemma, it follows 
that the condition (3) is not satisfied by $r$ wrt $S=W_k$ and $X$,
contradiction.
\end{enumerate}
\end{proof}
%
%
\begin{lemma}\label{lem:unfounded:greatest}
Let $S$ be a model state of disjunctive program $P$ such that
$S\cap X=\emptyset$.

Then $P$ has the greatest unfounded set ${\cal U}_P(S)$.
\end{lemma}
%
\begin{proof}{lem:unfounded:greatest}
It suffices to prove that the union $U$ of a class ${\cal C}$ of unfounded sets
for $P$ wrt $S$ is also an unfounded set for $P$ wrt $S$.

For any $a\in U$ and any $r\in P$ such that $a\in\head{r}$, there is an unfounded
set $X\in{\cal C}$ with $a\in X$.
Then there are three possibilities:
\begin{enumerate}
\item $\body{r}$ is false wrt $S$;

\item There exists $x\in X$ such that $x\in\pbody{r}$:
It is obvious that $x\in U$;

\item $S\models\body{r}$ implies $S\models(\head{r}-X)$:

If $S\models\body{r}$, then there is a sub-disjunction $A$ of $\head{r}-X$
such that $A\in S$.

Since $U\cap S=\emptyset$, we have that $A\in (\head{r}-U)$.

This means that $U$ is also an unfounded set for $P$ wrt $S$.
\end{enumerate}
\end{proof}
Since we consider only finite propositional programs in this paper,
there is some $t\ge 0$ such that $W_t=W_{t+1}$.

Having Proposition~\ref{pro:unfounded:exi}, we can define
our disjunctive well-founded semantics $\uwfs$ in term of the
operator ${\mathcal W}$.
%
\begin{definition}
The well-founded semantics $\uwfs$ is defined by
\[
{\uwfs}(P)= \lfp({\mathcal W}_P).
\]
\end{definition}
%
For the program $P$ in Example~\ref{exa:dix},
${\uwfs}(P)=\{a\vee b, \naf c\}$.

An important result is that
$\wfdsnew$ is equivalent to $\uwfs$.
This means $\wfdsnew$ and $\dwfsnew$ can also be equivalently characterized
in term of the unfounded sets defined in this section.
%
\begin{theorem}\label{thm:unfounded}
For any disjunctive program $P$,
\[
{\wfds}(P)={\uwfs}(P).
\]
\end{theorem}
%
Theorem~\ref{thm:unfounded} provides further evidence
for suitability of WFDS (equivalently, $\dwfsnew$) as
the intended well-founded semantics for disjunctive
logic programs.

By the following lemma, we can directly prove Theorem~\ref{thm:unfounded}.
%
\begin{lemma}\label{lem:ws}
Let $P$ be a disjunctive program. Then
$W_k=S_k$ for any $k\ge 0$.
\end{lemma}
%
This lemma also reveals a kind of correspondence between the well-founded 
disjunctive hypotheses and the unfounded sets.

\begin{proof}{lem:ws}
We use induction on $k$:
it is obvious that $S_0=W_0=\emptyset$.
Suppose that $S_k=W_k$, we want to show that
$S_{k+1}=W_{k+1}$. By induction assumption, it suffices to show that
${\mathcal A}_P(S_k^-)=not{\mathcal U}_P(W_k)$. 
This is equivalent to prove that
$X_k={\mathcal U}_P(S_k)$
where $X_k=\{p\D \naf p\in {\mathcal A}_P(S_k^-)\}$. 
We prove this statement by the following two steps.
\begin{enumerate}
\item $X_k$ is an unfounded set of $P$ wrt $S_k$:

Assume that there is a rule $r$ in $P$:
$a_1\vee\cdots\vee a_n\la body(r)$ such that
neither the condition 1 nor 2 in Definition~\ref{def:unfounded}
is satisfied by $r$. 

Without loss of generality, assume that
$\{a_1,\ldots,a_u\}\subseteq X_k$ but
$\{a_{u+1},\ldots,a_n\}\cap X_k=\emptyset$. 

If $S_k\models body(r)$, 
then $\Delta'\leadsto_P\{\naf a_1,\ldots, \naf a_u\}$
where $\Delta'=S_k^-\cup\{\naf a_{u+1},\ldots,\naf a_n\}$.

Notice that $\{\naf a_1,\ldots,\naf a_u\}\subseteq {\mathcal A}_P(S_k^-)$,
it follows that $S_k^-\leadsto_P\Delta'$.
Since the hypothesis $S_k^-$ is self-consistent, it should be the
case that $\Delta'\leadsto_P \{\naf a_{u+1},\ldots,\naf a_n\}$.

This means that $S_k^-\vdash_P a_{i_1}\vee\cdots\vee a_{i_m}$
for a subset $\{a_{i_1},\cdots, a_{i_m}\}$ of $\{a_{u+1},\ldots,a_n\}$. 

Thus, $S_k\models a_{u+1}\vee\ldots\vee a_n$.
That is, $S_k\models (head(r)-X_k)$.

\item $X_k$ is the greatest unfounded set of $P$ wrt $S_k$:

We want to prove that each unfounded set $X$ of $P$ wrt $S_k$ is
a subset of $X_k$. 

It suffices to show that, for any $a\in X$, $\naf a$ is acceptable
by ${\mathcal A}_P(S_k^-)$. 

Notice that, for any $a\in X$,
the set $\{a\}$ is also an unfounded set of $P$ wrt $S_k$.

For any hypothesis $\Delta'$ such that $\Delta'\leadsto_P \{\naf a\}$,
then there is a rule $r$ of $P$:
\[
a\vee a_1\vee\cdots\vee a_n\la body(r)
\]
such that $\naf a_i\in\Delta'$ for $i=1,\ldots,n$ and $S_k^-\vdash_P body(r)$.

On the other hand, since $\{a\}$ is unfounded, we have that
$a_1\vee\cdots\vee a_n\in S_k$.

This implies that 
$S_k^-\leadsto_P\{\naf a_1,\ldots,\naf a_n\}$. That is,
$S_k^-\leadsto_P \Delta'$. 

Therefore, 
$\naf a \in {\mathcal A}_P(S_k^-)$.
\end{enumerate}  
\end{proof}
 
\section{Conclusion}\label{section:conclusion}
In this paper we have investigated recent
approaches to defining well-founded semantics for
disjunctive logic programs. 
We first provided a minor modification of the argumentative semantics
$\wfds$ defined in~\cite{wan00}.
Based on some intuitive program transformations,
we proposed an extension $\dwfsnew$ to the $\dwfs$ in~\cite{bradix99} by introducing 
a new program transformation
called the Elimination of s-implications. This transformation intuitively extends 
Brass and Dix's two program transformations (Elimination of nonminimal rules and
Negative reduction).
We have also given a new definition of the unfounded sets
for disjunctive programs, which is a generalization of the unfounded sets
investigated by \cite{eilesa97,lerusc97}.
This new notion of unfounded sets fully takes disjunctive information into
consideration and provides another interesting characterization for disjunctive
well-founded semantics. 
The main contribution of this paper is the equivalence of $\uwfs, \dwfs$ and
$\wfds$. We have also provided a bottom-up computation for these equivalent semantics.
A top-down procedure $\dsls$ is presented in~\cite{wan01b}, which is sound and complete
with respect to these three semantics. Therefore, the results shown
in this paper together with that in~\cite{wan01b} show that the following 
disjunctive well-founded semantics are equivalent:
\begin{itemize}
\item $\dwfsnew$ based on program transformation;
\item $\wfds$ based on argumentation;
\item $\uwfs$ based on unfounded sets;
\item $\dsls$ based on resolution.
\end{itemize}
These results show that, despite diverse proposals on defining disjunctive well-founded
semantics, some agreements still exist. 
The fact that different starting points lead to the same 
semantics provides a strong support for $\wfds$ (equivalently, $\dwfsnew$, $\uwfs$ and
$\dsls$). 
However, it is unclear to us whether these equivalent semantics can be characterized
by modifying STATIC~\cite{prz95}.

\paragraph{Acknowledgments}
The author would like to thank Philippe Besnard, Alexander Bochman,
James Delgrande, Norman Foo, Thomas Linke, 
Torsten Schaub and Yan Zhang for helpful comments on this work. 
This work was supported by DFG under grant FOR~375/1-1,~TP~C.
NSFC under grant 69883008.
\bibliographystyle{plain}

\end{document}